\documentclass[letterpaper]{article} 
\usepackage{aaai25}  
\usepackage{times}  
\usepackage{helvet}  
\usepackage{courier}  
\usepackage[hyphens]{url}  
\usepackage{graphicx} 
\usepackage{natbib}  
\usepackage{caption} 
\usepackage{algorithm}
\usepackage{colortbl}
\usepackage{booktabs}
\usepackage{algorithmicx}
\usepackage{algpseudocode}
\usepackage{amssymb}
\usepackage{amsmath}
\usepackage{booktabs}
\usepackage{comment}
\usepackage{ntheorem}
\usepackage{multirow}
\usepackage[]{tcolorbox}
\usepackage[shortlabels]{enumitem}
\newtheorem{definition}{Definition}
\newtheorem{lemma}{Lemma}

\algdef{SE}[DOWHILE]{Do}{doWhile}{\algorithmicdo}[1]{\algorithmicwhile\ #1}%

\theoremstyle{empty}

\tcbset{
    colback = white,
    before skip = 0.2cm,    
    after skip = 0.5cm      
}                           
\definecolor{my-blue}{cmyk}{0.56, 0.07, 0, 0.44, 1.00} 

\newtcolorbox{boxA}{
    boxrule = 1pt,
    colframe = my-blue,
    colback=white!10!white
}

\newtcolorbox{boxB}{
    boxrule = 0.2pt,
    colframe = black,
    colback=white!4!white
}
\usepackage{pifont}
\definecolor{mygreen}{RGB}{92, 214, 92}
\definecolor{myred}{RGB}{255, 0, 0}
\usepackage{newfloat}
\usepackage{listings}
\DeclareCaptionStyle{ruled}{labelfont=normalfont,labelsep=colon,strut=off} 
\floatstyle{ruled}
\newfloat{listing}{tb}{lst}{}
\floatname{listing}{Listing}
\title{RAZOR: Sharpening Knowledge by Cutting Bias with Unsupervised Text Rewriting}
\author{
    Shuo Yang\equalcontrib, Bardh Prenkaj\equalcontrib, Gjergji Kasneci\\
}
\affiliations{
    Technical University of Munich\\

    name.surname@tum.de
}

\usepackage{bibentry}

\begin{document}

\maketitle

\begin{abstract}
Despite the widespread use of LLMs due to their superior performance in various tasks, their high computational costs often lead potential users to opt for the pretraining-finetuning pipeline. However, biases prevalent in manually constructed datasets can introduce spurious correlations between tokens and labels, creating so-called \textit{shortcuts} and hindering the generalizability of fine-tuned models. Existing debiasing methods often rely on prior knowledge of specific dataset biases, which is challenging to acquire a priori. We propose RAZOR (Rewriting And Zero-bias Optimization Refinement), a novel, unsupervised, and data-focused debiasing approach based on text rewriting for shortcut mitigation. RAZOR leverages LLMs to iteratively rewrite potentially biased text segments by replacing them with heuristically selected alternatives in a shortcut space defined by token statistics and positional information. This process aims to align surface-level text features more closely with diverse label distributions, thereby promoting the learning of genuine linguistic patterns. Compared with unsupervised SoTA models, RAZOR improves by $3.5\%$  on the FEVER and $6.5\%$ on MNLI and SNLI datasets according to the F1 score. Additionally, RAZOR effectively mitigates specific known biases, reducing bias-related terms by $\times 2$ without requiring prior bias information, a result that is on par with SoTA models that leverage prior information. Our work prioritizes data manipulation over architectural modifications, emphasizing the pivotal role of data quality in enhancing model performance and fairness. This research contributes to developing more robust evaluation benchmarks for debiasing methods by incorporating metrics for bias reduction and overall model efficacy.
\end{abstract}

%

\section{Introduction}\label{sec:introduction}
With the progression of instructional learning, prominent LLMs such as ChatGPT~\citep{10.5555/3600270.3602281} and Vicuna~\citep{Zheng2023JudgingLW} are extensively utilized across diverse domains owing to their versatility and outstanding performance~\citep{10.1145/3641289}. However, their utility is significantly constrained by the high training and API licensing costs. As a result, numerous researchers in domains like healthcare, education, and e-commerce adhere to the conventional approach of using open-source language models followed by fine-tuning~\citep{strubell-etal-2019-energy}.

\begin{figure}[!t]
    \centering
        \begin{boxA}
            \small
            \textcolor{black}{
            \textbf{Spielberg} is a great spinner of a yarn, \underline{however} this time he \underline{just didn't do it} for me. ($f_\theta$: \textcolor{mygreen}{\checkmark}, $\Phi$: \textcolor{myred}{$\times$})
            \rule{\textwidth}{0.2pt}
            The benefits of a \textbf{New York Subway} system are that a person can get from A to B \underline{without being stuck in traffic} and subway trains are faster than buses. ($f_\theta$: \textcolor{myred}{$\times$}, $\Phi$: \textcolor{mygreen}{\checkmark})
            }
    \end{boxA}
    \caption{Example of spurious correlation in sentiment classification tasks, where a classifier $f_\theta$ takes \textit{Spielberg} and \textit{New York Subway} as shortcuts and makes wrong predictions w.r.t. the ground truth ($\Phi$). The classifier concentrates on the bold tokens to make the prediction; however, the underlined tokens might be more useful in producing the correct label.}
    \label{fig:spurious_correlation_example}
\end{figure}

LLMs absorb linguistic features from extensive corpora during pre-training by predicting specific lexical units. This foundational learning phase is then followed by fine-tuning domain-specific data, enhancing their ability to perform domain-specific tasks effectively. 

Unlike linguistic features, surface biases are reported as commonly existing in manually curated datasets and can detrimentally affect the fine-tuning phase~\citep{JIMENEZ2020102919}. In other words, models may achieve untruthfully elevated performance during training by exploiting specific shortcuts. For instance, analyses of the FEVER dataset~\citep{Thorne18Fever}, used for fact verification tasks, revealed a pronounced association between the occurrence of negations within the claim and the corresponding label ``refutes''~\citep{karimi-mahabadi-etal-2020-end}. This suggests that models might erroneously generalize that claims containing negations are invariably fake. 

The contemporary debiasing methods are relatively effective but require prior knowledge of biases. For example, recent approaches like CLEVER~\citep{xu-etal-2023-counterfactual} stand on the prior knowledge that shortcuts originate solely from tokens of specific parts within the input. Therefore, they can estimate and correct the impact of shortcuts. However, such prior knowledge is scarce and difficult to obtain. Furthermore, bias types vary significantly with changes in datasets~\cite {Geirhos2020ShortcutLI}, severely limiting the generality of contemporary methods based on supervised learning.

As an unsupervised improvement strategy, our rationale lies in the limitations of LLMs to explicitly identify specific tokens that serve as shortcuts~\cite {85d395469caa4cc383259d3012f528bc}. Consequently, we propose equalizing all token-related surface\footnote{Surface features do not contain any semantic/linguistic characteristics in a sentence.} features across samples from the different classes. By achieving balance among these features, models are prevented from leveraging their potentially spurious correlations with labels. This approach thereby compels deeper semantic learning and augments overall robustness.

The main contributions of this paper are as follows:\footnote{Our code is available at \url{https://github.com/ShuoYangtum/RAZOR}. The appendix sections can be found at \url{https://arxiv.org/abs/2412.07675}.}

\begin{itemize}
    \item We introduce RAZOR -- short for \textbf{R}ewriting \textbf{A}nd \textbf{Z}ero-bias \textbf{O}ptimization \textbf{R}efinement -- a novel unsupervised debiasing technique that mitigates shortcuts in NLP models through iterative text rewriting.

    \item To the best of our knowledge, we are the first to formalize the definition of shortcuts about an underlying classifier and the ground truth of the sentences. Our theoretical findings help us understand RAZOR's effectiveness in mitigating spurious correlations in real-world datasets without prior knowledge.
    
    \item  RAZOR improves over the SoTA, achieving a 3.5\% performance increase on the FEVER dataset and a 6.5\% boost on MNLI and SNLI datasets. 
    
    \item RAZOR reduces bias-related terms by $\times$$2$, matching the performance of SoTA models without requiring prior knowledge of bias. This makes RAZOR a crucial advancement for tackling bias in NLP, particularly in scenarios where explicit bias information is inaccessible.
    
    \item  Our approach emphasizes the understanding of shortcuts and highlights the power of data-driven techniques over architectural changes in bias mitigation. This reinforces the critical role of data quality in enhancing model fairness and performance while establishing a foundation for developing stronger evaluation benchmarks in future debiasing research.
\end{itemize}

\section{Related Work}
In recent years, significant research efforts have been directed toward identifying and mitigating spurious correlations in NLP tasks. Our work strictly relates to \textit{creating balanced and less biased datasets} and \textit{counterfactual data generation} lines of research. 

\paragraph{Creating Balanced and Less Biased Datasets.}

\citet{wu-etal-2022-generating} present a data generation approach aimed at mitigating spurious correlations in natural language inference datasets through diverse data generation. The authors fine-tune GPT-2 as their data generation model to sequentially engender an instance's premise, label, and hypothesis. Additionally, they apply unlikelihood training to mitigate label inconsistency phenomena, improving the label quality at generation time. Finally, they add a consistency step \cite{bartolo-etal-2021-improving,10.1162/tacl_a_00415} -- i.e., using the model's confidence for filtering purposes -- to improve the quality of the generated dataset further. After this process, the authors reject samples that contribute to the high spurious correlations between task-independent features of the samples and their labels via z-filtering and produce a debiased dataset. \citet{pmlr-v119-bras20a} propose an iterative greedy
algorithm that adversarially filters out spurious biases from
data by relying on AFLite \cite{DBLP:conf/aaai/SakaguchiBBC20}. Interestingly, the authors aim to mitigate performance leakage when datasets present spurious biases.\footnote{Spurious biases in datasets make benchmarks artificially easier as models learn to overly rely on these biases instead of learning to generalize. This also hurts out-of-domain generalization.} CrossAug \citep{DBLP:conf/cikm/LeeWKLPJ21} reduces dataset bias by creating contrastive pairs in fact-checking tasks. It uses a two-stage pipeline where the first stage generates negative claims by altering positive claims, and the second stage modifies the evidence to support these negative claims. In this way, CrossAug enhances a classification model's ability to reason more accurately based on the provided evidence piece of text. Finally, CLEVER \cite{xu-etal-2023-counterfactual} leverages counterfactual inference for debiasing in fact-checking tasks. Instead of relying on data augmentation or training-phase adjustments, CLEVER focuses on the inference stage. It does so by training two models, a claim-evidence fusion model, and a claim-only model, and then subtracting the output of the claim-only model from the fusion model to isolate and remove biases. EDA
\cite{DBLP:conf/emnlp/WeiZ19} involves four simple techniques: synonym replacement, random insertion, random swap, and random deletion, all modifying the text while preserving its original meaning. These operations generate diverse training examples, helping models generalize better by exposing them to varied linguistic patterns. The approach is designed to be easy to implement and computationally inexpensive. EDA mitigates overfitting and improves model robustness by focusing on random, controlled transformations. \citet{DBLP:conf/emnlp/SchusterSYFSB19} demonstrate that models trained solely on claims (without evidence) can achieve high accuracy due to dataset biases, particularly the presence of giveaway phrases that correlate with specific labels. To mitigate this issue, they create a ``Symmetric Test Set'', which eliminates these biases by manually generating balanced claim-evidence pairs. Additionally, they propose a regularization method -- namely ReW -- that reweights training examples to reduce the influence of these biases during model training. \citet{DBLP:conf/acl/MahabadiBH20} propose ``Product of Experts'' -- namely PoE -- which combines predictions from a bias-only model and the main model, focusing the training on examples that are not biased.

Differently from the above works, we defer the rewriting process to GPT-3.5-Turbo, which has demonstrated advantages to mere heuristic-based methods that change synonyms \cite{DBLP:conf/emnlp/WeiZ19}, negate verbs to engender new sentences \cite{DBLP:conf/cikm/LeeWKLPJ21}, or reweigh the sentences to downgrade biases \cite{DBLP:conf/emnlp/SchusterSYFSB19}. Similarly to \cite{wu-etal-2022-generating}, we use a consistency mechanism. However, we give GPT-3.5-Turbo the original evidence, the newly generated claim, and ask it to tell whether it supports/refutes the new claim and repeat the generation process until the class distributions are aligned (see Sec. \ref{sec:rewriting}).

\paragraph{Counterfactual Data Generation.} Lately, augmenting models with counterfactual data has become a favored method for reducing spurious correlations and enhancing model robustness. \citet{kaushik-lipton-2018-much} initially employed human workers to generate counterfactual examples for augmentation. Their findings indicate that counterfactually augmented data effectively mitigates spurious patterns in training data. However, this approach is costly, time-intensive, and often results in simple perturbations. \citet{wu-etal-2021-polyjuice} and \citet{ross-etal-2022-tailor} leverage text generation models to produce counterfactual data. These models require fine-tuning with specific perturbation types and have various limitations, including the untargeted and unlabelled generation process and the restricted perturbation types. To introduce new perturbation types, the models need retraining. Unlike the previous two methods,  \citet{chen-etal-2023-disco} develop DISCO, a framework for distilling counterfactuals with large language models to help distinguish between spurious and genuine correlations. DISCO does prompt engineering to generate phrasal perturbations with a large general language model. Then, a task-specific teacher model filters these generations to distill high-quality counterfactual data to enrich the original training set. Lastly, \citet{xu2023counterfactual} focus on counterfactual debiasing for fact verification, generating counterfactual scenarios to identify and mitigate spurious correlations in verification tasks.

\section{Method}
\subsection{Problem Formulation}
We consider binary classifications of short text documents, e.g., sentiment classification of reviews. Let us assume that we have a set of i.i.d. labelled samples -- i.e., dataset -- $\mathcal{D} = \{d_1,\dots,d_n\}$. Each sample in $\mathcal{D}$ is a document, i.e., a sequence of tokens,  $d_i = \{t_1,t_2,\dots,t_m\}$ with a corresponding label $y_i \in \mathcal{Y}$ where w.l.o.g. $\mathcal{Y} = \{0,1\}$. To classify a document $d_i$, we first transform it into a feature vector $\mathbf{x}_i$ via a designated transformation function $g: \mathcal{D} \rightarrow \mathcal{X}\subseteq\mathbb{R}^u$, where $\mathcal{X}$ is the domain of features. Then, $\mathbf{x}_i$ is classified by a classification function $f_\theta: \mathcal{X} \rightarrow \mathcal{Y}$ with model parameters $\theta$.  The parameters $\theta$ are typically estimated from $\mathcal{D}$ by optmizing a certain loss function $\theta^* \leftarrow \arg\min_\theta \mathcal{L}(\mathcal{D}, \theta)$.

Generally, the classification function $f_\theta$ adopted in NLP is a deep learning model (see among others \cite{naseem2020transformer,li2024graph}). Alas, as noticed in \cite{izmailov2022feature}, deep classifiers are known to rely on spurious features -- i.e., patterns that are correlated with the target on the training data but not inherently relevant to the learning problem, such as the image backgrounds when classifying the foreground. In NLP problems, such as sentiment classification tasks, certain tokens (e.g., words) can be associated with positive/negative classes due to their co-occurrence with the labels in the training data. Inspired by \cite{DBLP:conf/naacl/WangSY022}, in Fig. \ref{fig:spurious_correlation_example}, we show how the classifier exploits the shortcut terms \textit{Spielberg} and \textit{New York Subway} to misclassify the sentences. While shortcut mitigation has been extensively studied, to the best of our knowledge, there is a lack of a uniform definition of what a shortcut represents. Hence, we provide the reader with a formal definition of shortcut tokens. Notice that each document $d_i$ gets tokenized into a sequence of tokens $\{t_1,t_2,\dots,t_m\}$, e.g., at a word-level, character-level, or sub-word level before being transformed into a feature vector $x_i$. Lastly, assume that $\Phi: \mathcal{X} \rightarrow \mathcal{Y}$ represents the ground truth function.

\begin{definition}\label{def:shortcut}
    Given a document of tokens $d_i = \{t_1,t_2,\dots,t_m\} \in \mathcal{D}$, a feature transformation function $g: \mathcal{D} \rightarrow \mathcal{X}\subseteq\mathbb{R}^u$, a classifier $f_\theta: \mathcal{X} \rightarrow \mathcal{Y}$, and a ground truth function $\Phi: \mathcal{X} \rightarrow \mathcal{Y}$, a set of tokens $\hat{d}_i \subset d_i$ is a \underline{shortcut} if the following conditions are satisfied:
    \begin{equation}\label{eq:equality}
        f_\theta\big(g(\hat{d}_i)\big) = f_\theta\big(g(d_i)\big)
    \end{equation}
    \begin{equation}\label{eq:inequality}
        f_\theta\big(g(d_i)\big) \neq \Phi\big(g(d_i)\big)
    \end{equation}
    \begin{equation}\label{eq:cardinality}
        |\hat{d}_i| \leq |d_i\backslash\hat{d}_i|
    \end{equation}
\end{definition}
For the sake of the argument, let us assume that $f_\theta$ is a transformer. We indicate with $h: \mathcal{X} \rightarrow \mathbb{R}^{\ell}$ a score-attribution function to each input feature.  Notice that the following derivation can be applied for any attribution function -- either post-hoc, such as SHAP \cite{DBLP:conf/nips/LundbergL17} or not. We leave the derivation under a generic attribution function for future work and focus on the attention mechanism of a transformer in this work.

\noindent\textbf{Attention mechanism.} Let $h: \mathcal{X} \rightarrow \mathbb{R}^\ell$ be the attention mechanism that, given in input a document $d_i = \{t_1,t_2,\dots,t_m\}$, produces attention weights $\mathbf{z}_j$ for each token $t_j \in d_i$. In this work, we assume that $h(g(t_j)) = \mathbf{z}_j\;\forall t_j \in d_i$.


\begin{figure*}[!t]
        \centering     \includegraphics[width=.8\linewidth]{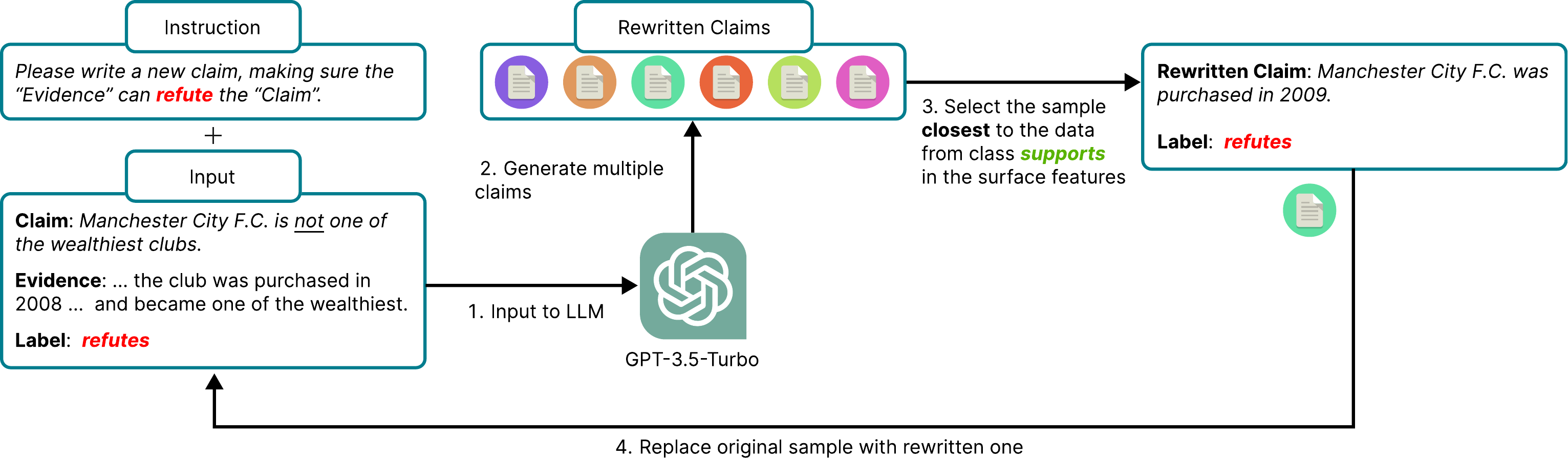}
            \caption{An example of RAZOR's application in the fact-checking task. The task aims to determine whether a piece of evidence from Wikipedia \textbf{\textcolor{mygreen}{supports}} or \textbf{\textcolor{myred}{refutes}} a claim. The instance here is sampled from the FEVER dataset, where a negation word "not" has been reported to exhibit a spurious correlation with the class \textbf{\textcolor{myred}{refutes}}.}
            \label{fig:example}
            \vspace{-1em}
\end{figure*}

\noindent\textbf{Essential vs. non-essential tokens.} According to Definition~\ref{def:shortcut}, we have two sets of tokens for each document $d_i$: i.e., the set of ``essential'' tokens $\hat{d}_i$, and that of ``non-essential'' ones $d_i\backslash \hat{d}_i$. The essential tokens are the ones that contribute the most (i.e., their attention scores are high) to the prediction. In contrast, non-essential tokens do not significantly contribute to the prediction \cite{vaswani2017attention}. Because the prediction does not change if we use the entire set of tokens in the document $d_i$, intuitively, one can conclude that the attention scores of the set of non-essential are lower than those of the essential set of tokens (see Lemma \ref{lemma:attention_scores}).

\begin{lemma}\label{lemma:attention_scores}
    
    Given a document of tokens $d_i = \{t_1,t_2,\dots,t_m\}\in\mathcal{D}$, and an attention scores attribution function $h: \mathcal{X} \rightarrow \mathbb{R}^\ell$, if $\hat{d}_i \subset d_i$ is a shortcut, then
    \begin{equation}\label{eq:attention_scores}
    \frac{H(\hat{d}_i)}{|\hat{d}_i|} \geq \frac{H(d_i\backslash\hat{d}_i)}{|d_i\backslash\hat{d_i}|}
    \end{equation}
    where
    \begin{equation}\label{eq:inequality_attention}
            H(\hat{d}_i) = \bigg|\bigg|\sum_{t_j \in \hat{d}_i} h\big(g(t_j)\big)\bigg|\bigg|_2
    \end{equation}
    \begin{equation}
           H(d_i\backslash\hat{d}_i) = \bigg|\bigg|\sum_{t_j \in d_i\backslash\hat{d}_i} h\big(g(t_j)\big)\bigg|\bigg|_2
    \end{equation}
\end{lemma}
We invite the reader to check Appendix A for the proof of Lemma \ref{lemma:attention_scores}. In practice, we observe that shortcuts are much smaller than the rest of the tokens in a sentence, which paves the way for efficient methods to identify the subset of tokens that represent shortcuts.

\subsection{Shortcut Identification}
Throughout the rest of the paper, we use document and sentence interchangeably. In this paper, we propose \textbf{R}ewriting \textbf{A}nd \textbf{Z}ero-bias \textbf{O}ptimization \textbf{R}efinement -- namely RAZOR -- a sentence rewriting strategy that mitigates shortcuts and promotes dataset debiasing. RAZOR aims to rewrite $\mathcal{D}$ by balancing the number of essential words between positive and negative sentences. However, the cost of rewriting all sentences in $\mathcal{D}$ is exceedingly high. As a heuristic optimization approach, we first filter a subset $\widehat{\mathcal{D}}$ that most likely includes shortcuts -- according to Eq. \ref{eq:shortcut_score} -- rewriting it and repeating this filter-rewriting process.

To find $\widehat{\mathcal{D}}$, we define a surface feature extraction function $\hat{g}: \mathcal{D} \rightarrow \mathbb{R}^\lambda$ which maps all samples into a shortcut space. Notably, the specific types of shortcuts are strongly task-dependent. In this work, without any prior knowledge, we provide a general identification pattern with the two essential factors by following Definition~\ref{def:shortcut}: i.e., \textbf{(1)} the \textbf{significance} of tokens, and \textbf{(2)} the (relative) \textbf{positions} of tokens. 

We first use the TF-IDF score to calculate the importance of a specific token $t_j \in d_i$. To achieve this, we define $S: \mathcal{D} \rightarrow \mathbb{R}$ as in Eq.~\eqref{eq:tf_idf}:
\begin{equation}\label{eq:tf_idf}
    S(t_j)=\frac{n(t_j,d_i)}{|d_i|} \log \frac{|\mathcal{D}|}{|\{d_k\;|\;\exists d_k \in \mathcal{D} \text{ s.t. } t_j \in d_k\}|}
\end{equation}%
where $n(t_j,d_i)$ is the number of occurrences of token $t_j$ in $d_i$. Applying Eq.~\eqref{eq:tf_idf} offers a key advantage: i.e., the TF-IDF-based measurement keeps the balance of token frequency, while word frequency is one of the factors most likely to cause shortcuts~\citep{85d395469caa4cc383259d3012f528bc}. For example, suppose a token holds considerably different frequencies in positive versus harmful data. As per how transformers work, this essential token will hold higher attention scores during training, leading the model $f_\theta$ to focus more on its occurrence rather than the semantic information of the sample -- i.e., this token is considered a shortcut.

Second, we inject positional information of words through the positional embeddings since the model $f_\theta$ may learn shortcuts from the relative positions of tokens. Specifically, we compare positional embeddings of the same token from samples with different labels to ensure they are closely aligned. Without prior knowledge, we use a fixed positional encoding function $\tau: \mathbb{N} \rightarrow \mathbb{R}^\lambda$ based on sine and cosine functions
\begin{equation}
    \tau(\text{pos}) = \Bigg[\text{PE}(\text{pos},0),\dots,\text{PE}(\text{pos},\lambda-1)\Bigg],
\end{equation}
with
\begin{equation}
\text{PE}(\text{pos},k) = 
\begin{cases}
      \text{sin}\bigg(\frac{\text{pos}}{10000^{2k/\lambda}}\bigg) & k \equiv 0 \;(\text{mod } 2) \\
       \text{cos}\bigg(\frac{\text{pos}}{10000^{2k/\lambda}}\bigg) & k \equiv 1\; (\text{mod } 2)
\end{cases}
    \label{eq:positional_embedding}
\end{equation}
where pos is the token's position in a target document, $k$ is the dimension index in the embedding vector, and $\lambda$ is the dimensionality of the positional embedding vector. Afterwards, we compute the mean of the products of $S$ for all tokens in a document $d_i$ as its shortcut feature mapping
\begin{equation}\label{eq:scalar_prod}
    \hat{g}(d_i)=\frac{\sum_{t_j \in d_i}{S(t_j) * \tau(j)}}{|d_i|-1},
\end{equation}
assuming that $t_j$ occurs in position $j$ in the document $d_i$, and $*$ is the scalar product. Finally, let us use $\vartheta(d_i, \Phi) $ to indicate
\begin{equation}\label{eq:diff_label_docs}
    \vartheta(d_i, \Phi) = \{ d_j \;|\; d_j\in \mathcal{D} \text{ s.t. } \Phi(d_i) \neq \Phi(d_j)\}.
\end{equation}
In other words, given a document $d_i \in \mathcal{D}$, and a ground truth function $\Phi$, as introduced in Definition \ref{def:shortcut}, $\vartheta(d_i,\Phi)$ produces a set of documents whose label is different with $\Phi(d_i)$. In this way, we define a \textit{shortcut score} $\gamma: \mathcal{D} \rightarrow \mathbb{R}$ as in Eq. \ref{eq:shortcut_score}.
\begin{equation}\label{eq:shortcut_score}
    \gamma(d_i) = 1 - \frac{1}{\big|\vartheta(d_i,\Phi)\big|}\sum_{d_j \in \vartheta(d_i,\Phi)} \underbrace{\frac{\hat{g}(d_i)\cdot\hat{g}{(d_j)}}{||\hat{g}(d_i)||\;||\hat{g}(d_j)||}}_{\text{cos}(\hat{g}(d_i),\hat{g}(d_j))}
\end{equation}%
Note that the higher the shortcut score, the greater the token position distribution differs between the sample $d_i$, assuming it is positive, and the negative samples in $\varphi(d_i,\Phi)$, indicating that the model $f_\theta$ is more likely to learn shortcuts from $d_i$. We select the top documents with the highest shortcut values to form $\widehat{\mathcal{D}}$ and use the documents therein for rewriting.

\subsection{Rewriting and Filtering}\label{sec:rewriting}
After obtaining the set of shortcut candidates, we rewrite each sentence $d_i \in \widehat{\mathcal{D}}$  into several candidate sentences maintaining the original label $\Phi(d_i)$. We then substitute $d_i$ with the rewritten sentence least likely to contain shortcuts. In Fig. \ref{fig:example}, we show how this rewriting process works in a fact-checking task. Here, we aim for the LLM to generate a new \textit{claim} based on the same \textit{evidence} from Wikipedia while ensuring that the evidence still supports (or refutes) the newly generated claim. To rewrite the sentence $d_i$, we construct a prompt for an LLM that generates rewritten candidates. The prompt contains two components: i.e., an \textit{instruction}, which is a general explanation of the rewriting task for $d_i$ including $\Phi(d_i)$; and the \textit{input} $d_i$ itself. We ask the LLM to engender rewritten sentences while maintaining the original label $\Phi(d_i)$. Hence, we rely on a self-consistency mechanism to ensure the quality of generated sentences. Specifically, in the fact-checking scenario, we provide the LLM with the original evidence and the newly generated claim and query it to tell us whether the evidence supports/refutes the new claim. More formally, let $G_\alpha: \mathcal{D} \times \mathcal{Y} \rightarrow \mathcal{D}$ be the LLM we are querying to generate new sentences respecting the original label. Additionally, let $G_\beta: \mathcal{D} \times \mathcal{Y} \rightarrow \{0,1\}$ be another LLM\footnote{Note that we exploit these LLMs in zero-shot. Therefore, $G_\alpha$ and $G_\beta$ can be the same kind of LLM -- e.g., GPT-4o -- as long as there are two different sessions between the generation and the verification processes.} that verifies whether the generated sentence has the original label. More formally, we accept an engendered sentence $d_i^* = G_\alpha(d_i,\Phi(d_i))$ if Eq. \eqref{eq:label_check}  is satisfied.
\begin{equation}\label{eq:label_check}
\mathbb{I}\bigg[G_\beta\bigg(G_\alpha\big(d_i,\Phi(d_i)\big),\Phi(d_i)\bigg) = 1\bigg]
\end{equation}%
Once, the rewriting process is completed for a particular $d_i$, we obtain a set of generated documents $\varphi(d_i) = \{d_i^*\;|\;d_i^* = G_\alpha(d_i,\Phi(d_i))\}$. From $\varphi(d_i)$, we choose the one scoring the least according to Eq. \ref{eq:shortcut_score} and insert it in the original dataset $\mathcal{D}$. The original document $d_i$ is removed from the ``rewritten'' dataset as we aim to mitigate the shortcuts that it might have contained. In this way, RAZOR maintains the same size as the original dataset throughout the rewriting process. We repeat this process until the cosine similarity between the embeddings in the surface feature space of the documents of each label is maximal (see Eq. \ref{eq:kl}).
\begin{equation}\label{eq:kl}
\begin{gathered}
    \text{maximize} \sum_{d_i \in \mathcal{D}_0} \sum_{d_j \in \mathcal{D}_1} \text{cos}(\hat{g}(d_i),\hat{g}(d_j))\\
    \text{s.t. } \mathcal{D}_y = \{d \;|\; d \in \mathcal{D}\;\land\;\Phi(d) = y\}
\end{gathered}
\end{equation}

\section{Experiments}

\begin{table*}[ht]
\centering
\caption{Comparison of RAZOR with SoTA methods on the FEVER and FEVER-Adversarial datasets -- the datasets have the same training set but different test sets. We show the performance of the classifiers in terms of accuracy, and F1 scores after the shortcuts within the dataset have been modified using each debiasing method, including RAZOR (ours). \textit{Original} means that the original training set has not been modified, and we report the performances of the two classifiers as is.}
\label{tab:fever}
\resizebox{.9\textwidth}{!}{%
\begin{tabular}{@{}c@{\hspace{.2\tabcolsep}}clcccccccc@{}}
\toprule
\multicolumn{3}{l}{\multirow{3}{*}{}} &
  \multicolumn{4}{c}{FEVER} &
  \multicolumn{4}{c}{FEVER-Adversarial} \\ \cmidrule(l){4-11} 
\multicolumn{3}{l}{} &
  \multicolumn{2}{c}{BERT} &
  \multicolumn{2}{c}{RoBERTa} &
  \multicolumn{2}{c}{BERT} &
  \multicolumn{2}{c}{RoBERTa} \\
\multicolumn{3}{l}{} &
  Accuracy &
  F1 &
  Accuracy &
  F1 &
  Accuracy &
  F1 &
  Accuracy &
  F1 \\ \midrule
\multicolumn{2}{r}{} &
  Original &
  87.93$\pm$0.51 &
  86.65$\pm$0.78 &
  90.05$\pm$0.48 &
  88.90$\pm$0.69 &
  64.10$\pm$1.21 &
  69.81$\pm$1.45 &
  66.99$\pm$1.30 &
  73.13$\pm$1.82 \\ \midrule
\multirow{7}{*}{\rotatebox{90}{Rewriting}} &
  \multirow{7}{*}{\rotatebox{90}{Strategy}} &
  EDA &
  87.99$\pm$0.30 &
  87.03$\pm$0.50 &
  89.25$\pm$0.32 &
  89.00$\pm$0.66 &
  65.20$\pm$0.99 &
  69.55$\pm$0.86 &
  66.35$\pm$1.54 &
  69.98$\pm$1.97 \\
 &
   &
  CrossAug  &
  \textbf{90.28$\pm$0.42} &
  \textbf{89.37$\pm$0.54} &
  \textbf{91.00$\pm$0.46} &
  \textbf{91.98$\pm$0.78} &
  63.58$\pm$1.20 &
  66.74$\pm$1.56 &
  67.02$\pm$1.44 &
  72.04$\pm$1.77 \\
 &
   &
  ReW  &
  87.50$\pm$0.88 &
  87.76$\pm$1.19 &
  89.09$\pm$1.21 &
  89.44$\pm$2.01 &
  64.52$\pm$2.21 &
  69.85$\pm$2.30 &
  67.23$\pm$2.03 &
  \underline{74.20$\pm$2.68} \\
 &
   &
  PoE &
  86.25$\pm$1.78 &
  86.98$\pm$1.89 &
  90.20$\pm$1.78 &
  89.79$\pm$1.90 &
  65.20$\pm$2.76 &
  \underline{70.45$\pm$2.58} &
  \underline{68.00$\pm$2.49} &
  72.12$\pm$2.61 \\
 &
   &
  CLEVER &
  85.13$\pm$1.20 &
  85.33$\pm$2.03 &
  86.55$\pm$1.72 &
  84.54$\pm$2.33 &
  56.16$\pm$2.54 &
  58.93$\pm$2.27 &
  62.63$\pm$2.96 &
  63.35$\pm$2.35 \\ 
 &
   &
  RAZOR (w/ LLaMA) &
  87.97$\pm$1.20 &
  87.22$\pm$2.03 &
  \underline{90.59$\pm$2.32} &
  89.11$\pm$2.56 &
  \underline{66.03$\pm$1.95} &
  69.61$\pm$1.88 &
  \underline{68.77$\pm$3.01} &
  73.76$\pm$2.50 \\ 
 &
   &
  RAZOR (w/ GPT) &
  \underline{88.73$\pm$2.23} &
  \underline{88.39$\pm$2.20} &
  90.45$\pm$1.99 &
  \underline{90.22$\pm$2.11} &
  \textbf{66.66$\pm$1.98} &
  \textbf{73.78$\pm$1.37} &
  \textbf{69.21$\pm$1.64} &
  \textbf{76.08$\pm$2.02} \\ \bottomrule
\end{tabular}%
}
\end{table*}

\begin{table*}[]
\centering
\caption{RAZOR's impact on the classifiers on the NLI datasets in terms of Accuracy and F1 score. Note that we assess the performances of the classifiers across different train-test set combinations -- e.g., training on MNLI and testing on SNLI -- to mitigate the fact that the original test set contains similar shortcuts as the original training set. Within parentheses, we indicate the training set; outside of them, we indicate the test set.}
\label{tab:NLI}
\resizebox{.9\textwidth}{!}{%
\begin{tabular}{@{}lccccccccc@{}}
\toprule
\multirow{2}{*}{} &
  \multicolumn{2}{c}{MNLI (MNLI)} &
  \multicolumn{2}{c}{SNLI (MNLI)} &
  \multicolumn{1}{l}{} &
  \multicolumn{2}{c}{MNLI (SNLI)} &
  \multicolumn{2}{c}{SNLI (SNLI)} \\ \cmidrule(l){2-10} 
 &
  Accuracy &
  F1 &
  Accuracy &
  F1 &
  \multicolumn{1}{l}{} &
  Accuracy &
  F1 &
  Accuracy &
  F1 \\ \midrule
BERT &
  86.70$\pm$1.21 &
  86.85$\pm$1.79 &
  83.05$\pm$1.38 &
  84.41$\pm$2.01 &
   &
  70.15$\pm$2.89 &
  62.75$\pm$2.77 &
  91.65$\pm$0.80 &
  91.74$\pm$0.91 \\
w/ RAZOR (GPT) &
  88.50$\pm$1.32 &
  87.90$\pm$1.80 &
  84.75$\pm$1.72 &
  85.07$\pm$2.44 &
   &
  74.10$\pm$3.20 &
  67.86$\pm$3.57 &
  93.20$\pm$1.11 &
  93.16$\pm$1.26 \\
w/ RAZOR (LLaMA) &
  86.80$\pm$1.32 &
  86.88$\pm$1.80 &
  84.10$\pm$1.52 &
  83.51$\pm$2.00 &
   &
  74.25$\pm$2.92 &
  68.73$\pm$3.14 &
  92.90$\pm$1.27 &
  92.81$\pm$1.32 \\ \cmidrule(l){2-10} 
$\Delta$ (GPT) &
  +2.08 &
  +1.44 &
  +2.04 &
  +0.78 &
   &
  +5.63 &
  +8.14 &
  +1.69 &
  +1.55 \\
$\Delta$ (LLaMA) &
  +0.12 &
  +0.04 &
  +1.26 &
  -1.06 &
   &
  +5.85 &
  +9.56 &
  +1.36 &
  +1.17 \\ \midrule\midrule
RoBERTa &
  90.85$\pm$1.10 &
  90.75$\pm$1.53 &
  90.30$\pm$1.27 &
  91.08$\pm$1.55 &
   &
  85.35$\pm$2.04 &
  84.93$\pm$1.48 &
  95.05$\pm$0.72 &
  95.01$\pm$0.80 \\
w/ RAZOR (GPT) &
  91.15$\pm$1.26 &
  91.14$\pm$1.48 &
  91.45$\pm$1.29 &
  91.42$\pm$1.68 &
   &
  86.50$\pm$2.20 &
  86.22$\pm$1.72 &
  95.15$\pm$0.70 &
  95.17$\pm$0.98 \\
w/ RAZOR (LLaMA) &
  91.55$\pm$1.38 &
  91.45$\pm$1.70 &
  90.90$\pm$1.29 &
  91.13$\pm$1.68 &
   &
  85.50$\pm$1.98 &
  85.76$\pm$2.24 &
  95.60$\pm$0.70 &
  95.71$\pm$0.98 \\ \cmidrule(l){2-10} 
$\Delta$ (GPT) &
  +0.33 &
  +0.42 &
  +1.27 &
  +0.37 &
   &
  +1.34 &
  +1.52 &
  +0.11 &
  +0.17 \\
$\Delta$ (LLaMA) &
  +0.77 &
  +0.77 &
  +0.66 &
  +0.05 &
   &
  +0.18 &
  +0.98 &
  +0.58 &
  +0.74 \\ \midrule\midrule
DistilBERT &
  83.50$\pm$1.23 &
  82.97$\pm$1.77 &
  77.23$\pm$1.85 &
  77.38$\pm$1.86 &
   &
  65.55$\pm$2.30 &
  69.90$\pm$1.78 &
  88.82$\pm$1.22 &
  87.78$\pm$1.67 \\
w/ RAZOR (GPT) &
  85.14$\pm$1.66 &
  85.20$\pm$1.84 &
  80.22$\pm$1.90 &
  78.82$\pm$2.21 &
   &
  68.58$\pm$2.16 &
  72.80$\pm$1.83 &
  91.82$\pm$0.88 &
  91.80$\pm$1.03 \\
w/ RAZOR (LLaMA) &
  84.00$\pm$1.35 &
  83.98$\pm$1.78 &
  77.35$\pm$1.43 &
  77.62$\pm$1.99 &
   &
  66.33$\pm$2.00 &
  71.30$\pm$1.85 &
  90.30$\pm$1.26 &
  90.23$\pm$1.97 \\ \cmidrule(l){2-10} 
$\Delta$ (GPT) &
  +1.96 &
  +2.69 &
  +3.87 &
  +1.86 &
   &
  +4.62 &
  +4.14 &
  +3.38 &
  +4.58 \\
$\Delta$ (LLaMA) &
  +0.60 &
  +1.22 &
  +0.16 &
  +0.31 &
   &
  +1.19 &
  +2.00 &
  +1.67 &
  +2.79 \\ \bottomrule
\end{tabular}%
}
\end{table*}




\paragraph{Datasets.}
Fact-checking tasks aim to determine the veracity of a claim from social media based on existing facts. Here, we use the FEVER dataset \citep{Thorne18Fever}, which consists of claims and evidence retrieved from Wikipedia. The training objective is to classify whether the evidence supports or refutes the claim. Due to the prevalence of generating non-factual claims by simply negating factual claims, a significant negation bias has been reported in the literature.

Natural Language Inference (NLI) aims to determine the semantic relationship between a pair of sentences. Given a premise and a hypothesis, the goal is to classify the relationship as entailment, contradiction, or neutral. Here, we rely on the Multi-Genre Natural Language Inference (MNLI) \cite{N18-1101} corpus and the Stanford Natural Language Inference (SNLI) \cite{DBLP:conf/emnlp/BowmanAPM15} corpus, which provides a large number of sentence pairs annotated with their logical relationship. 

\paragraph{Experimental setup.} Our preprocessing steps involve truncating sequences to 512 tokens. We rely on three classification models to measure RAZOR's effectiveness, namely BERT, RoBERTa, and DistilBERT -- taken from HuggingFace\footnote{For BERT, we use the \texttt{bert-base-uncased} pre-trained model. For RoBERTa, see \url{https://huggingface.co/FacebookAI/roberta-base}. For DistilBERT, see \url{https://huggingface.co/distilbert/distilbert-base-uncased}.} -- followed by a linear classification layer. We train the classifiers using the LoRa algorithm \cite{DBLP:conf/iclr/HuSWALWWC22} and the AdamW optimizer with a batch size of $16$ and a learning rate of $3 \times 10^{-5}$. In the sentence rewriting, we set $G_\alpha = \text{GPT-3.5-Turbo}$ and $G_\beta = \text{GPT-3.5-Turbo}$. Here, we control the diversity of generated sequences by setting the top-p value to $0.9$ and the temperature to $0.7$. To provide details on the impact of the selected LLM to rewrite the sentences and verify their labels, we change both $G_\alpha$ and $G_\beta$ to LLaMA-3.1-8B-Instruct (see Tables~\ref{tab:fever} and~\ref{tab:NLI}).

\subsection{Discussion}

\paragraph{RAZOR is a more robust dataset debiaser than SoTA methods on fact-checking tasks with, respectively, $\mathbf{\sim}$$\mathbf{2\%}$ and $\mathbf{\sim}$$\mathbf{3.63\%}$ increase in Accuracy and F1 over the second-best.} We train BERT, RoBERTa, and DistilBERT classifiers on the original FEVER dataset to represent our baseline (hereafter Original). Then, for each SoTA method, we train the two classifiers on the modified version of FEVER according to the proposed algorithms. Notice that RAZOR and other SoTA methods only rewrite the training set of the dataset. The test set remains unchanged. We follow this strategy to assess the impact of the rewriting strategy w.r.t. the test set's original peculiarities, which might still contain shortcuts. We expect that if the rewriting strategy truly alleviates the shortcut problem in the training set, then in the test set, we will see an increase in performance since the false negative/positive ratio should decrease\footnote{As per Definition \ref{def:shortcut}, a shortcut causes classifiers to have false positives/negatives.}. Table \ref{tab:fever} shows this effect on the two classifiers mentioned above for RAZOR and SoTA methods. Notice how, although second-best after CrossAug, RAZOR effectively removes the shortcut effect on both classifiers w.r.t. the original baseline. We acknowledge that RAZOR is behind CrossAug in terms of performance because the latter enriches the dataset with contrastive -- i.e., positive-negative -- claim-evidence pairs, where the evidence is rewritten differently from RAZOR, which only operates at the claim level. However, we show how RAZOR is more robust than CrossAug in the FEVER-Adversarial dataset.\footnote{The training set of FEVER and FEVER-Adversarial is the same. The test sets are different.} The sentences in the test set of FEVER-Adversarial do not contain shortcuts. Hence, we use the adversarial test set to demonstrate RAZOR's debiasing abilities. Notice how CrossAug's performance is systematically lower than that of RAZOR. Because CrossAug negates the claim and evidence to generate other factual instances, it might happen that this newly generated claim-evidence pair can itself contain shortcuts, exacerbating the classifier's reasoning capabilities. However, since FEVER's test set contains shortcuts, CrossAug's original performances are not faithful to assess the goodness of this debiasing strategy. In particular, this is evident in FEVER-Adversarial, where simple negations of claim-evidence pairs are not optimal. Contrarily, RAZOR reduces shortcuts' effect and achieves consistently better results than all SoTA methods.

\paragraph{RAZOR demonstrates cross-dataset generalizability in multi-class classification tasks.}
To validate the generalizability of our method, we rely on MNLI and SNLI for a multi-class classification test. Similarly to the above process, we fine-tuned the classifiers as baselines to determine whether the logical relationship between two sentences is entailment, contradiction, or neutral. Subsequently, we used RAZOR to rewrite the training sets and achieve improvements in Accuracy and F1 scores for the two classifiers. With the lack of corresponding attacking datasets like FEVER-Adversarial, we exchanged MNLI and SNLI test sets to simulate a real-world data distribution. 

Our experiments demonstrate that even widely used datasets exhibit shortcut problems. We observed that by applying RAZOR, the baseline classifiers achieved improvements of approximately 0.1 to 8 points across all datasets and metrics. The most significant improvement is observed in the SNLI-MNLI training-test set combination. This suggests that SNLI may contain more shortcuts than MNLI, resulting in a performance difference of over 20 points for BERT across different test sets. However, after applying RAZOR, a substantial portion of the shortcuts in the data was removed, leading to an increase of 8.14 points in F1 score and 5.63 points in accuracy for BERT. Moreover, we find that RAZOR performs better on all classifiers overall. This is because models with better overall performances are less susceptible to shortcut issues. Therefore, RAZOR demonstrates significant potential for debiasing smaller models.

\paragraph{RAZOR effectively reduces the number of negations considered shortcuts in FEVER by 36.2\%.} 
\begin{figure}[ht]
        \centering     \includegraphics[width=\linewidth]{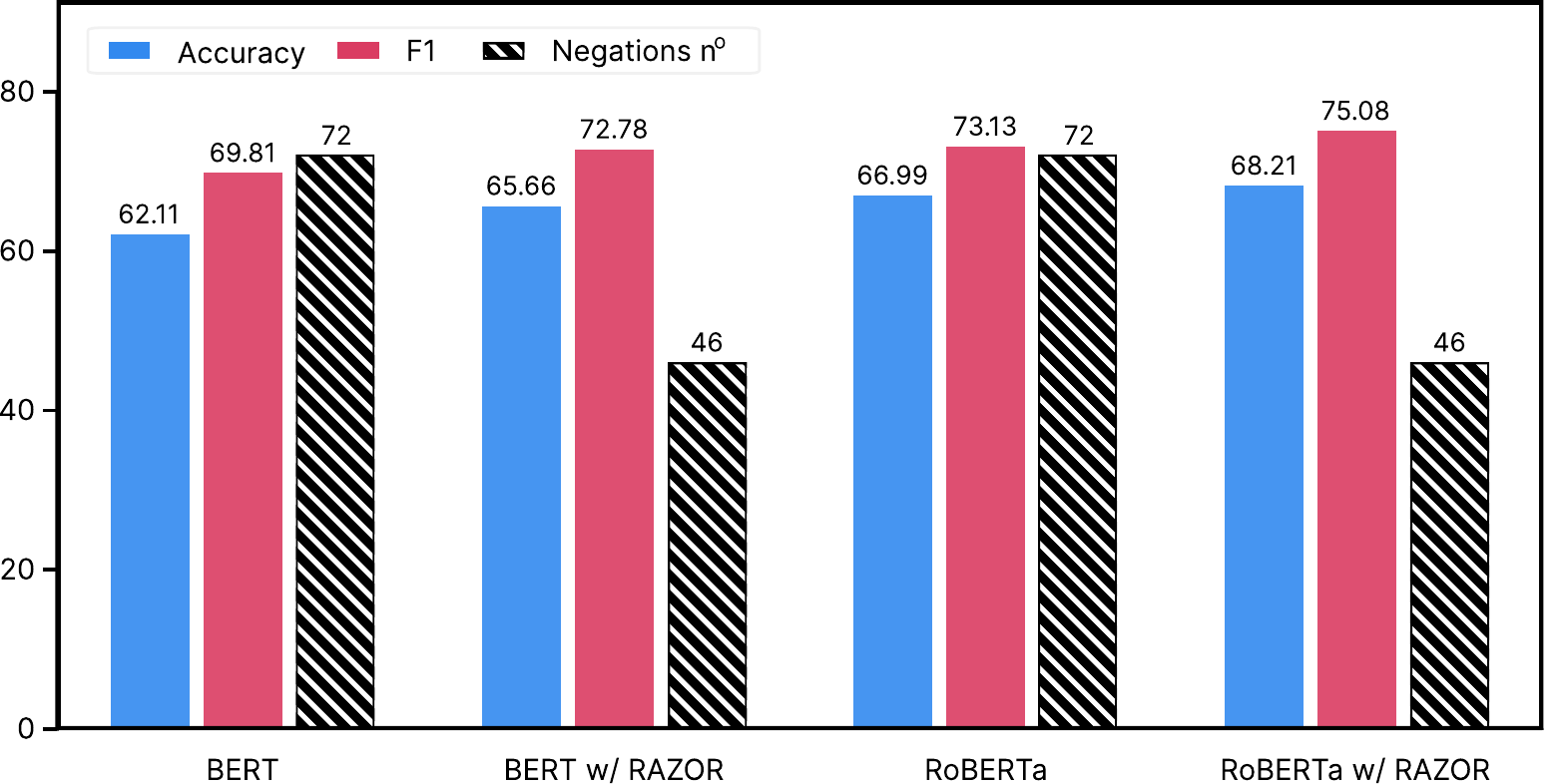}
            \caption{Effect of shortcut-related terms for BERT and RoBERTa with and without RAZOR on 500 randomly sampled original-rewritten pairs on the FEVER dataset. We then test on the FEVER-Adversarial set.}
            \label{fig:negations}
\end{figure}

To further quantify the effectiveness of our approach in mitigating shortcuts, we perform a statistical analysis of shortcut-related terms, specifically negations, within both the original and rewritten FEVER training set. Following the methodology presented in \cite{li-etal-2024-prompting}, we train models with randomly sample 500 original-rewritten pairs from the two sets, count the occurrences of the words "no" and "not" in their claims, and repeat the experiments detailed in Table~\ref{tab:fever} for the test set of the FEVER-Adversarial dataset. We illustrate the results in Fig.~\ref{fig:negations}. Notice that, by applying RAZOR, the occurrence of shortcut-related terms is reduced by approximately 40\% without any prior knowledge of these terms. Furthermore, eliminating shortcuts significantly improved models' generalizability, leading to an increase of, respectively, $\sim$$3$ and $\sim$$2$ percentage points for BERT and RoBERTa in both Accuracy and F1 scores when we test on the FEVER-Adversarial set.

\subsection{Ablation Studies}

\begin{figure}[!t]
        \centering     \includegraphics[width=\linewidth]{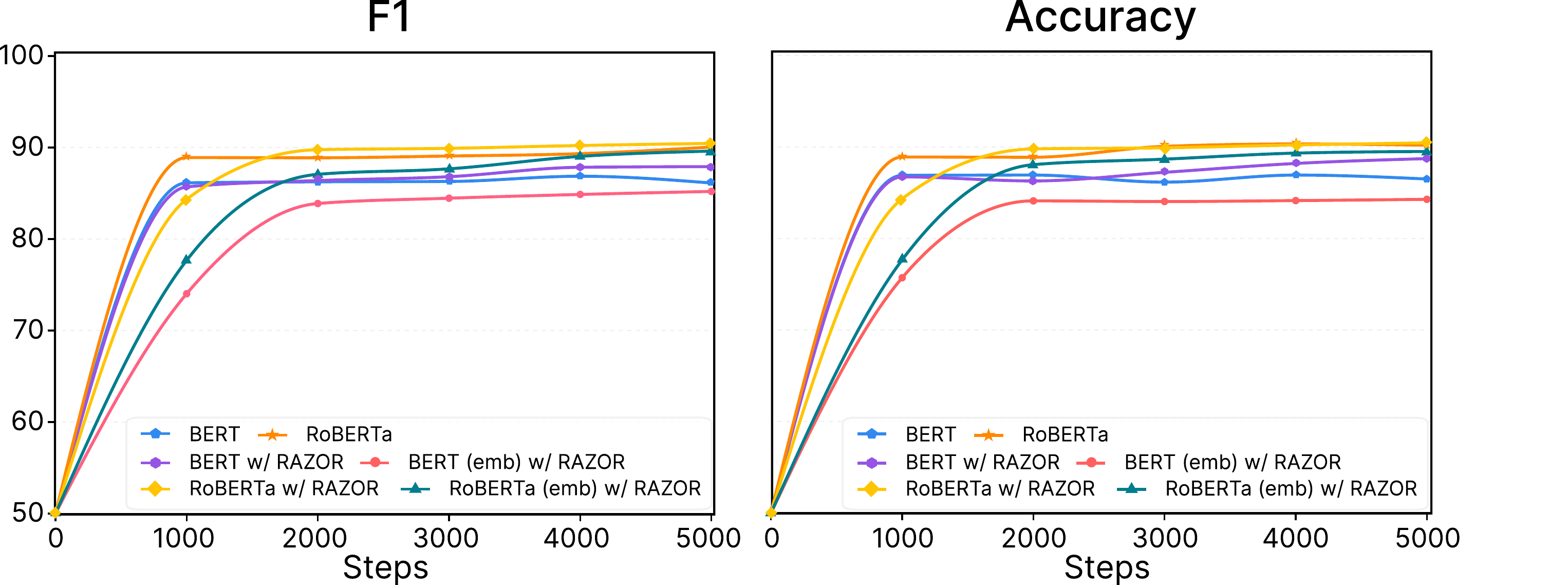}
            \caption{The performance of models trained on FEVER when evaluated on the original FEVER test set.}
            \vspace{-10pt}
            \label{fig:ablation}
\end{figure}

To ensure that the performance improvement of our model does not stem from GPT itself, we first ruled out potential information leakage by calculating repetition rates. Subsequently, we replaced our shortcut space mapping with the model's token embeddings and employed the same method to demonstrate the effectiveness of our defined shortcuts. We illustrate the results in Fig.~\ref{fig:ablation}. We observe consistent trends in the Accuracy and F1 score training curves. RAZOR outperforms the baseline and significantly surpasses the sentence embedding-based method on RoBERTa and BERT.

Furthermore, RAZOR exhibits a slightly slower convergence rate than the baseline but meets equilibrium earlier than the embedding-based method. We attribute this to RAZOR's ability to mitigate the influence of surface features, encouraging the classifiers to move beyond specific token biases and invest more time in learning linguistic features, ultimately resulting in superior performance. In contrast, embedding-based methods rely on sentence representations that mix surface and linguistic features, failing to separate the two effectively, preventing the classifiers from avoiding shortcuts. Moreover, the proximity of embeddings may obscure the classification hyperplane, inherently complicating the task and leading to mediocre performance. Given that both methods employ GPT-3.5-Turbo but produce different outcomes, we can safely conclude that the overall RAZOR's framework, rather than solely GPT in the rewriting process, is a primary driver in its effectiveness. Appendix B illustrates the performance difference between RAZOR operating on BERT with surface features and embedding features.

\section{Conclusion}
RAZOR is an effective dataset debiasing method. It outperforms SoTA methods in addressing shortcut problems in fact-checking and natural language inference tasks. RAZOR maintains dataset integrity by generating new sentences that adhere to the original labels and strategically removing the original documents while significantly reducing shortcut-induced biases. This is achieved through an iterative rewriting process, optimized until the KL divergence between the embeddings in the surface feature space is minimized. Our experiments consistently show that RAZOR enhances the performance of classifiers such as BERT and RoBERTa, particularly in challenging adversarial scenarios where traditional methods like CrossAug fail to deliver competitive results. Notably, RAZOR's superior performance on models like BERT, even compared to more robust models like RoBERTa, underscores its potential as a powerful debiasing tool, particularly for smaller and more resource-constrained models. The implications of our work extend beyond the datasets and models tested, offering a promising approach to improving model reliability across a wide range of machine-learning applications. While RAZOR has shown great promise, future work could explore its application to other types of biases or in conjunction with additional debiasing strategies. We believe that RAZOR represents a significant step forward in the ongoing effort to create more fair and accurate machine learning systems, ultimately contributing to more trustworthy AI applications in the real world.

\bibliography{aaai25}

\begin{appendix}
\section{Proof of Lemma 1}\label{app:proof}

During training, the model adjusts its parameters to minimize a loss function $\mathcal{L}$, which measures the difference between the model's predictions and the true labels. The attention scores $\mathbf{z}_j = h(g(t_j))$ for each token $t_j \in d_i$ in the sequence are influenced by this optimization process. Because of Eq. (1), the $\hat{d}_i$ information is sufficient for $f_\theta$ to decide. Now, the attention mechanism in transformers is designed to assign higher weights to more relevant tokens. Given that $\hat{d}_i$ contains sufficient information for the prediction, we can safely assume that

\begin{equation}\label{eq:expected_values_inequality}\tag{15}
    \mathbb{E}\bigg[\big|\big|h\big(g(t_j)\big)\big|\big|_2\;|\; t_j \in \hat{d}_i\bigg] \geq \mathbb{E}\bigg[\big|\big|h\big(g(t_j)\big)\big|\big|_2\;|\; t_j \in d_i\backslash\hat{d}_i\bigg].
\end{equation}%
Now, let us consider the left side of the inequality of Eq. 4 -- see also Eq. 5 -- by the triangle inequality, we have
\begin{equation}\label{eq:shortcuts}\tag{16}
\begin{gathered}
    H(\hat{d}_i) = \bigg|\bigg|\sum_{t_j \in \hat{d}_i} h\big(g(t_j)\big)\bigg|\bigg|_2 \leq \sum_{t_j \in \hat{d}_i} \bigg|\bigg|h\big(g(t_j)\big)\bigg|\bigg|_2.
\end{gathered}
\end{equation} 
Similarly, for the right side, we have
\begin{equation}\label{eq:nonshortcuts}\tag{17}
\begin{gathered}
    H(d_i\backslash\hat{d}_i) = \bigg|\bigg|\sum_{t_j \in d_i\backslash\hat{d}_i} h\big(g(t_j)\big)\bigg|\bigg|_2 \leq \sum_{t_j \in d_i\backslash\hat{d}_i} \bigg|\bigg|h\big(g(t_j)\big)\bigg|\bigg|_2.
\end{gathered}
\end{equation}%
If we consider the expected values (i.e., average norms per token), the inequalities in Eqns.~\eqref{eq:shortcuts} and~\eqref{eq:nonshortcuts} can be rewritten as follows
\begin{equation}\tag{18}
    H(\hat{d}_i) \leq  \big|\hat{d}_i\big| \cdot \mathbb{E}\bigg[\big|\big|h\big(g(t_j)\big)\big|\big|_2\;|\; t_j \in \hat{d}_i\bigg],
\end{equation}
\begin{equation}\label{eq:nonshortcut_expected}\tag{19}
    H(d_i\backslash\hat{d}_i) \leq \big|d_i\backslash \hat{d}_i\big|\cdot\mathbb{E}\bigg[\big|\big|h\big(g(t_j)\big)\big|\big|_2\;|\; t_j \in d_i\backslash\hat{d}_i\bigg].
\end{equation}
Because of Eqns.~\eqref{eq:expected_values_inequality} -- \eqref{eq:nonshortcut_expected}, we can state that
\begin{equation}\tag{20}
    H(\hat{d}_i) \geq H(d_i\backslash\hat{d}_i).
\end{equation}
And, since $|\hat{d}_i| \leq |d_i\backslash\hat{d}_i|$, as per Eq. (3), then Eq. (4) holds. $\square$

\section{Additional Experiments}\label{app:add_experiments}

\paragraph{RAZOR effectively separates the classes on FEVER-Adversarial.}

\begin{figure}[!ht]
        \centering     \includegraphics[width=\linewidth]{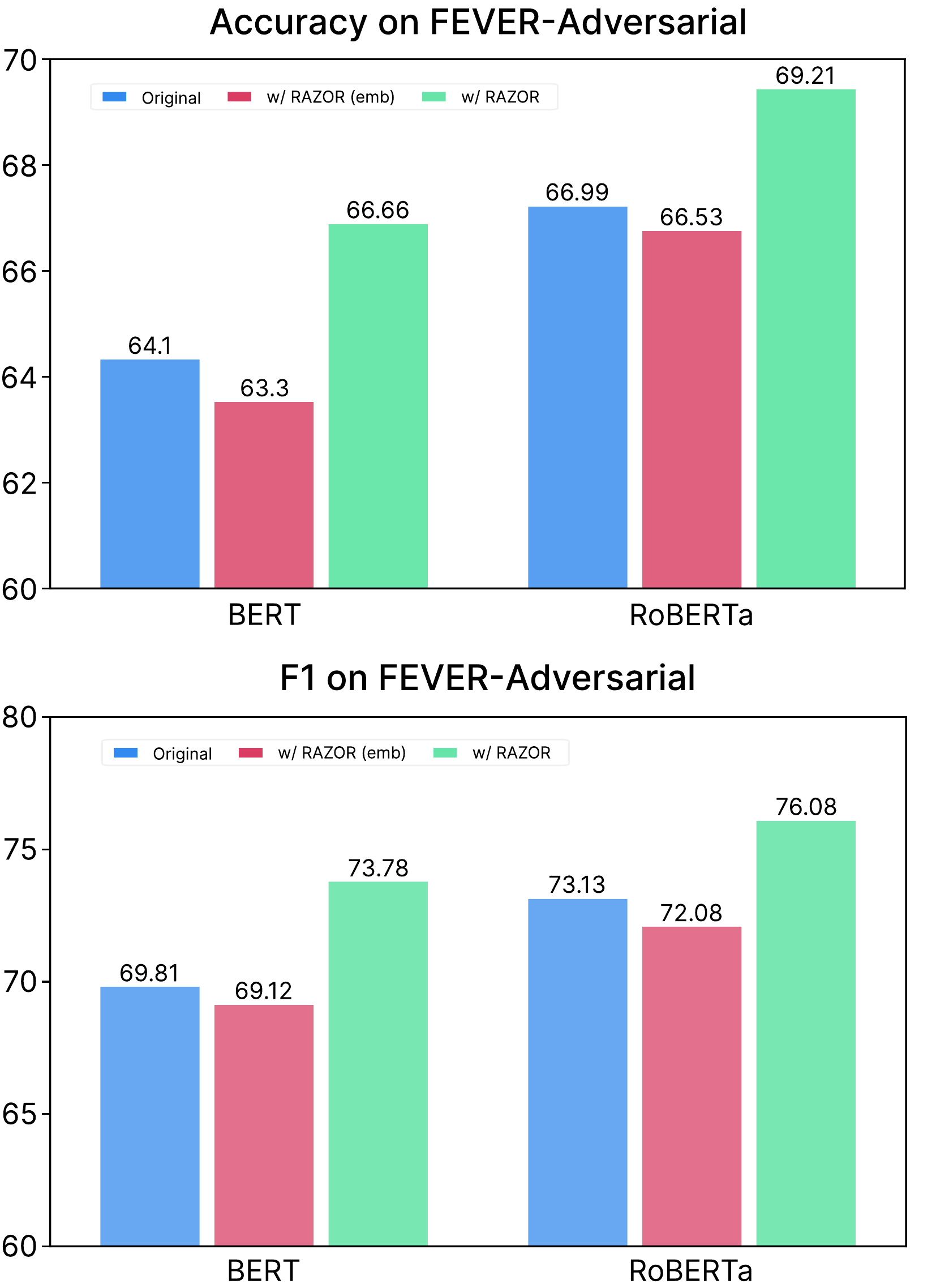}
            \caption{The performance of models trained on 500 samples randomly selected from FEVER when evaluated on the FEVER-Adverasial test set.}
            \label{fig:ablation_adversarial}
\end{figure}

\begin{figure*}[!ht]
        \centering     \includegraphics[width=.7\textwidth]{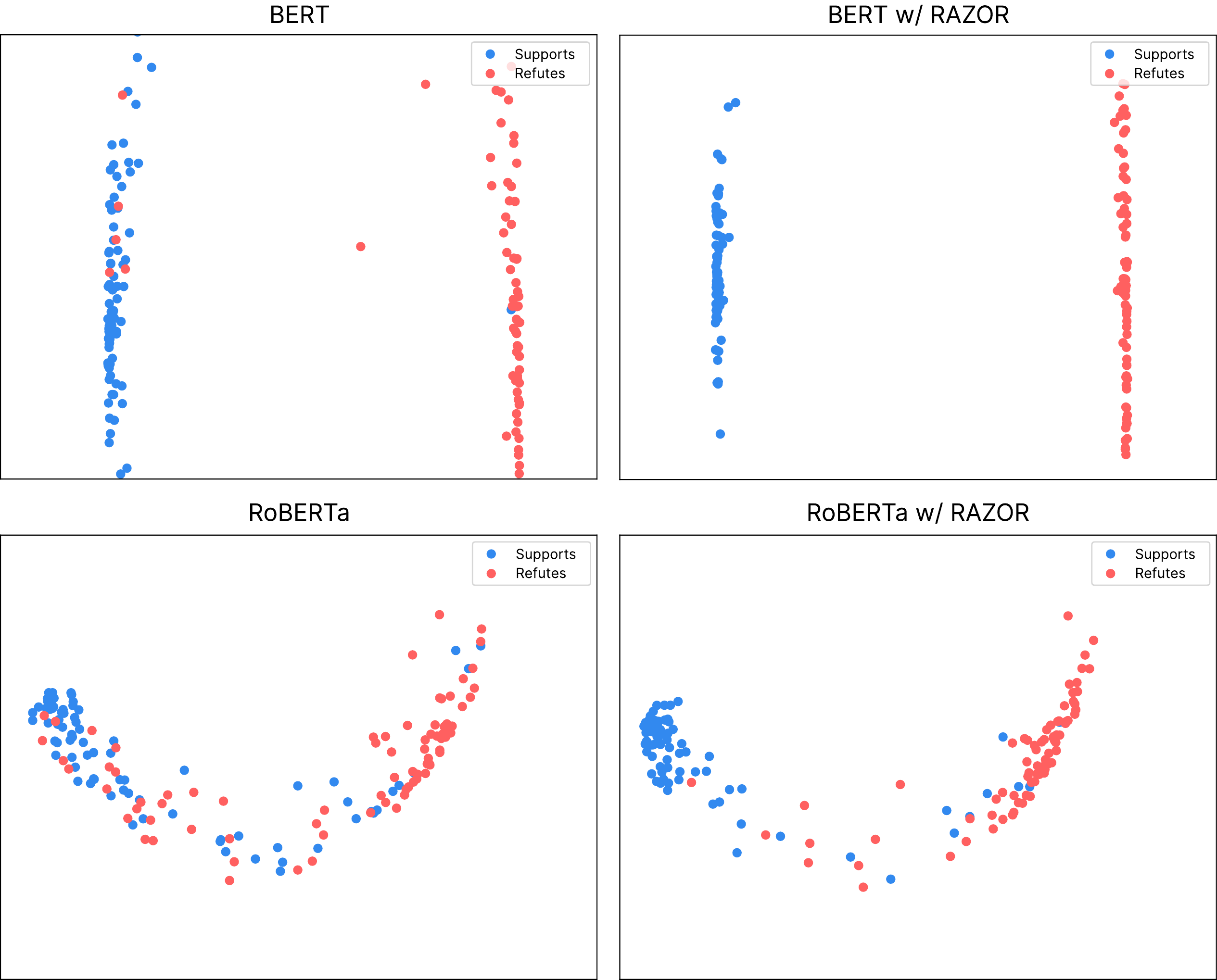}
            \caption{PCA visualization of FEVER-Adversarial (test) samples, comparing logits from BERT and RoBERTa fine-tuned on the original and the rewritten training sets with RAZOR.}
            \label{fig:PCA}
\end{figure*}

In Fig. \ref{fig:PCA}, we illustrate RAZOR's effect on the test samples of the FEVER-Adversarial dataset on BERT and RoBERTA. Notice how RAZOR helps reduce class confusion, especially for RoBERTa.

\paragraph{RAZOR does not require large datasets to remove shortcuts in fact-checking datasets correctly.}
Fig. \ref{fig:ablation_adversarial} presents additional results for our ablation study. We observed that under a small data size, RAZOR still yields noticeable improvements of 2-4 points in Accuracy and F1 score. Furthermore, compared to the embedding-based RAZOR, we analyze that the improvement is attributed solely to eliminating surface shortcuts rather than GPT-3.5-Turbo.

\begin{table}[ht]
\centering
\caption{Corpus-Level BLEU scores for the original and rewritten training sets evaluated on two test sets.}
\label{tab:bleu}
\resizebox{\linewidth}{!}{%
\begin{tabular}{lcc}
\toprule
         & \multicolumn{1}{c}{FEVER} & \multicolumn{1}{c}{FEVER-Adversarial} \\ \midrule
Original & 13.33                     & 11.02                           \\
w/ RAZOR   & 12.02                     & 22.41                           \\ \bottomrule
\end{tabular}
}

\end{table}

\paragraph{RAZOR effectively mitigates lexical shortcuts.} Table~\ref{tab:bleu} presents the corpus-level BLEU scores~\citep{papineni-etal-2002-bleu} for the original training set and the rewritten ones with RAZOR, evaluated on the tests sets of FEVER and FEVER-Adversarial. After rewriting, the BLEU scores between the original data and the Adversarial test set, which is more generalizable, increased twofold, indicating that RAZOR effectively mitigates biases at the surface feature level.

\section{User-Evaluation Survey and Examples for Explanations on Fact-Checking Datasets}\label{app:user_eval}

\begin{table}[ht]
\centering
\caption{User evaluation results. ``REA.'' is reasonableness. ``COM.'' is comprehensiveness.}
\label{tab:user_eval}
\resizebox{\linewidth}{!}{%
\begin{tabular}{lccc}
\toprule
              & REA.        & COM.        & \begin{tabular}[c]{@{}c@{}}Explanation Preference Ratio\\ (Orig. vs. Rewritten Claim)\end{tabular} \\ \midrule
RoBERTa       & 2.50 / 7.00 & 2.24 / 7.00 & 21\%                                                                                               \\
w/ RAZOR & 4.14 / 7.00 & 3.70 / 7.00 & 79\%                                                                                               \\ \bottomrule
\end{tabular}%
}
\end{table}

Fact verification means utilizing Wikipedia as evidence to evaluate user-generated claims online, determining whether factual evidence supports or refuses these claims. We conducted a user-study evaluation test\footnote{We had 10 participants in the survey.} on ten randomly selected evidence samples in the FEVER dataset. Each question contained the original claim-evidence pair and the corresponding rewritten claim with RAZOR for the same evidence. The tokens on the evidence for the original and rewritten claim are colored according to SHAP -- i.e., the bluer the token, the more support for the claim; the redder, the more the refusal. The participants were asked to rate the rationality and comprehensiveness of the explanations on the evidence for the original and rewritten claims on a scale of 1 to 7. The participants were also asked to choose their preference between the explanation of the original claim and the rewritten claim. Here, we want to demonstrate that the debiased model -- in this case, RoBERTa w/ RAZOR -- is less influenced by shortcuts compared to the original model, and, thus, RAZOR aids in generating more reasonable classification ``justifications''. Table \ref{tab:user_eval} illustrates the results of this study. Note that the rewritten claims contain better explanations with a margin of 1.64 and 1.46 in, respectively, reasonableness and comprehensiveness. Moreover, we observe a 79\% preference for the explanations on the evidence for the rewritten claim with both refusal and support tokens -- i.e., RoBERTa w/ RAZOR - from the participants, leading us to believe that human-centered evaluations add-on to the usefulness of RAZOR besides it having SoTA performances (see Tables 1 and 2).

Here, we present one piece of evidence from Wikipedia along with two related claims. Different words within the evidence potentially impacting the reasoning process are highlighted in red and blue. The task of the participants was to assess whether these highlights were appropriate.
\begin{itemize}
    \item The red text represents key points that refute the corresponding claim. The deeper the red, the less likely the claim will be true based on this part of the text.
    \item The blue text highlights key points that support the corresponding claim. The deeper the blue, the more likely the claim is true based on this part of the text.
\end{itemize}
For each question, the participants were asked to answer the following regarding an original claim and a rewritten one: 
\begin{enumerate}
    \item Can the words marked in red and blue reasonably explain why the corresponding claim is supported or rejected?
    \item Have all the words that could serve as reasons for explanation been highlighted?
\end{enumerate}

\begin{figure}[!ht]
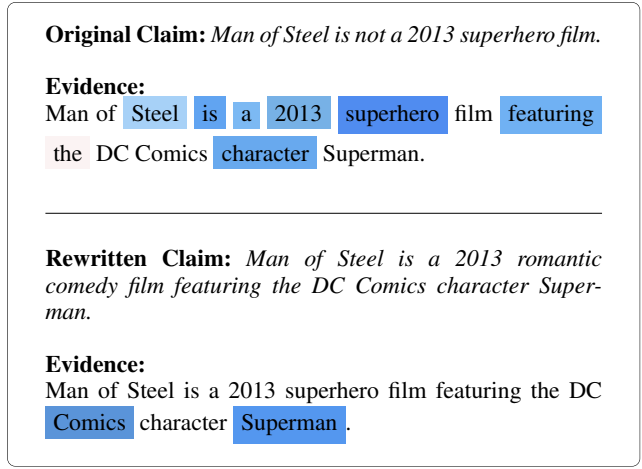

    \centering
        \begin{boxB}
            \small
            \textcolor{black}{
            \textbf{Original Claim:} \textit{Man of Steel is not a 2013 superhero film.}\\\\
            \textbf{Evidence:}\\
            Man of \colorbox[HTML]{A7D1F7}{Steel} \colorbox[HTML]{61A4F1}{is} \colorbox[HTML]{7CB9F3}{a} \colorbox[HTML]{77B1E6}{2013} \colorbox[HTML]{508CF0}{superhero} film \colorbox[HTML]{70B1F3}{featuring} \colorbox[HTML]{FBF3F3}{the} DC Comics \colorbox[HTML]{68A9EE}{character} Superman.\\ \\
            \rule{\textwidth}{0.2pt}\\\\
            \textbf{Rewritten Claim:} \textit{Man of Steel is a 2013 romantic comedy film featuring the DC Comics character Superman.}\\\\
            \textbf{Evidence:}\\
            Man of Steel is a 2013 superhero film featuring the DC \colorbox[HTML]{5A94D9}{Comics} character \colorbox[HTML]{5497EF}{Superman}.
            }
    \end{boxB}
    \caption{Example of claim-evidence pair with SHAP values explaining RoBERTa's classification.}
    \label{fig:user_eval}
\end{figure}

In Fig. \ref{fig:user_eval}, we show an example of a claim-evidence pair with the SHAP values explaining RoBERTa's classification (support/refusal). Then, we show the rewritten claim -- i.e., RoBERTa w/ RAZOR -- and the new SHAP explanation on the evidence for this claim. Here, for the original claim, the evidence refutes it because ``\textit{Man of Steel}'' is indeed a superhero film that came out in 2013. However, the text ``\textit{is a 2013 superhero}'' is wrongly highlighted with the color blue, which means these words support the claim instead of being highlighted with red. Hence, we would expect a lower user evaluation score for the question ``\textit{For the original claim, is the explanation for the highlighted words reasonable?}''. Regarding the rewritten claim, the text ``\textit{is a 2013 superhero}'' should be a reason to refute the claim. However, it is not highlighted. Therefore, we would expect a lower user evaluation score for the question ``\textit{For the rewritten claim, have all words that could serve as an explanation been marked?}''

\newpage
\section{RAZOR Algorithm}
Algorithm~\ref{al:razor} illustrates the entire process of how RAZOR operates. In the rewriting process -- in particular, line 13 -- we assume that it implicitly checks for consistency (see Eq. (13)). Therefore, $\varphi(d_i)$ here only contains rewritten documents that agree with the label $\Phi(d_i)$.
\begin{algorithm}[!h]
        \caption{\texttt{RAZOR}: shortcut identification, rewriting and filtering procedure.}
        \label{al:inter}        \begin{algorithmic}[1]
        \Require $\mathcal{D}=\{d_{1}, ..., d_{n}\}$ s.t. $d_{i} = \{t_1,...,t_m\}$, $k$, $\Phi: \mathcal{D} \rightarrow \mathcal{Y}$, $G_\alpha: \mathcal{D} \rightarrow \mathcal{D}$, $G_\beta: \mathcal{D} \rightarrow \mathcal{D}$, $\hat{g}: \mathcal{D} \rightarrow \mathbb{R}$, $\vartheta: \mathcal{D} \times \Phi \rightarrow \mathcal{D}$, $\gamma: \mathcal{D} \rightarrow \mathbb{R}$, $\varphi: \mathcal{D} \times \Phi \rightarrow \mathcal{D}$
        \Do
        \State $s \gets \emptyset$
        \For{$d_i \in \mathcal{D}$}
            \State Compute $\hat{g}(d_i)$ as in Eq. (10)
            \State Compute $\vartheta(d_i, \Phi)$ as in Eq. (11)
            \State $x \gets \gamma(d_i)$ according to Eq. (12)
            \State $s \gets s \cup \{x\}$ 
        \EndFor
            \State $\widehat{\mathcal{D}} \gets \text{\texttt{sort}}(\mathcal{D},s, \text{\texttt{desc}})$ \quad\textit{/* sort $\mathcal{D}$ w.r.t. $s$ */}
            \State $\mathcal{D}^\prime \gets \mathcal{D}[$:$k]$ \quad\textit{/* choose top-$k$ */}
           \For{$d_i \in \mathcal{D}^\prime$}
                \State $s \gets \emptyset$

                \State $\varphi(d_i) = \{d_i^*\;|\;d_i^* = G_\alpha(d_i,\Phi(d_i))\}$
                \For{$d_i^* \in \varphi(d_i)$}
                    \State $x \gets \gamma(d_i^*)$ according to Eq. (12)
                    \State $s \gets s \cup \{x\}$
                \EndFor

                \State $\hat{d}_i \gets \min s$
                \State Replace $d_i$ with $\hat{d_i}$ in $\widehat{\mathcal{D}}$.
            \EndFor
            \doWhile{$$\sum_{d_i \in \mathcal{D}_0} \sum_{d_j \in \mathcal{D}_1} cos(\hat{g}(d_i),\hat{g}(d_j))$$ is maximized where $\mathcal{D}_y = \{d \;|\; d \in \mathcal{D}\;\land\;\Phi(d) = y\}$}
            \State \Return $\widehat{\mathcal{D}}$
        \end{algorithmic}
        \label{al:razor}
    \end{algorithm}

\end{appendix}
\end{document}